\documentclass[lettersize,journal]{IEEEtran}
\usepackage{amsmath,amsfonts}
\usepackage{algorithmic}
\usepackage{algorithm}
\usepackage{array}
\usepackage{textcomp}
\usepackage{stfloats}
\usepackage{url}
\usepackage{verbatim}
\usepackage{graphicx}
\usepackage{cite}
\usepackage{enumitem}
\usepackage[mathscr]{eucal}
\usepackage[caption=false,font=scriptsize,labelfont=rm,textfont=rm]{subfig}
\usepackage{booktabs}
\usepackage{float}   
\usepackage{subfloat}
\usepackage{ragged2e} 
\usepackage{bbm}
\usepackage{bm}
\usepackage{xcolor}
\usepackage{hyperref}
\hypersetup{hypertex=true,
	colorlinks=true,
	linkcolor=blue,
	anchorcolor=blue,
	citecolor=blue}
\usepackage{orcidlink}
\usepackage{tikz}
\usepackage{multirow}
\usepackage{multicol}
\begin{document}

\title{Source-free Domain Adaptive Object Detection in Remote Sensing Images}

\author{Weixing Liu\hspace{-1.0mm}$^{~\orcidlink{0000-0002-0681-5257}}$,  Jun Liu\hspace{-1.0mm}$^{~\orcidlink{0000-0002-8943-079X}}$, Xin Su\hspace{-1.0mm}$^{~\orcidlink{0000-0003-0957-4628}}$, Han Nie\hspace{-1.0mm}$^{~\orcidlink{0000-0001-9229-6109}}$, and Bin Luo\hspace{-1.0mm}$^{~\orcidlink{0000-0002-3040-3500}}$,~\IEEEmembership{Senior Member,~IEEE} 

\thanks{Weixing Liu, Jun Liu, Han Nie, and Bin Luo are with the State Key Laboratory of Information Engineering in Surveying, Mapping and Remote Sensing, Wuhan University, Wuhan 430079, China.

Xin Su is with School of Remote Sensing and Information Engineering, Wuhan University, Wuhan 430079, China}}
\markboth{Journal of \LaTeX\ Class Files,~Vol.~14, No.~8, August~2021}%
{Shell \MakeLowercase{\textit{et al.}}: A Sample Article Using IEEEtran.cls for IEEE Journals}
\maketitle

\begin{abstract}
Recent studies have used unsupervised domain adaptive object detection (UDAOD) methods to bridge the domain gap in remote sensing (RS) images. However, UDAOD methods typically assume that the source domain data can be accessed during the domain adaptation process. This setting is often impractical in the real world due to RS data privacy and transmission difficulty. To address this challenge, we propose a practical source-free object detection (SFOD) setting for RS images, which aims to perform target domain adaptation using only the source pre-trained model. We propose a new SFOD method for RS images consisting of two parts: perturbed domain generation and alignment. The proposed multilevel perturbation constructs the perturbed domain in a simple yet efficient form by perturbing the domain-variant features at the image level and feature level according to the color and style bias. The proposed multilevel alignment calculates feature and label consistency between the perturbed domain and the target domain across the teacher-student network, and introduces the distillation of feature prototype to mitigate the noise of pseudo-labels. By requiring the detector to be consistent in the perturbed domain and the target domain, the detector is forced to focus on domain-invariant features. Extensive results of three synthetic-to-real experiments and three cross-sensor experiments have validated the effectiveness of our method which does not require access to source domain RS images. Furthermore, experiments on computer vision datasets show that our method can be extended to other fields as well. Our code will be available at: https://weix-liu.github.io/.
\end{abstract}


\section{Introduction}
\IEEEPARstart {R} {emote} sensing (RS) image object detection is an important research and application topic in RS image interpretation \cite{cheng2016learning}. In recent years, the rapid development of deep learning has significantly improved the performance of object detection in RS images \cite{fu2020rotation,li2020object,liu2021abnet,ma2022feature}. However, RS image object detection models based on deep learning require a large amount of expensive annotated data to complete training, such as UCAS-AOD\cite{zhu2015orientation}, xView\cite{lam2018xview}, and DIOR dataset\cite{li2020object}.

   \begin{figure}[!t]
    \centering
    \includegraphics[width=0.95\linewidth]{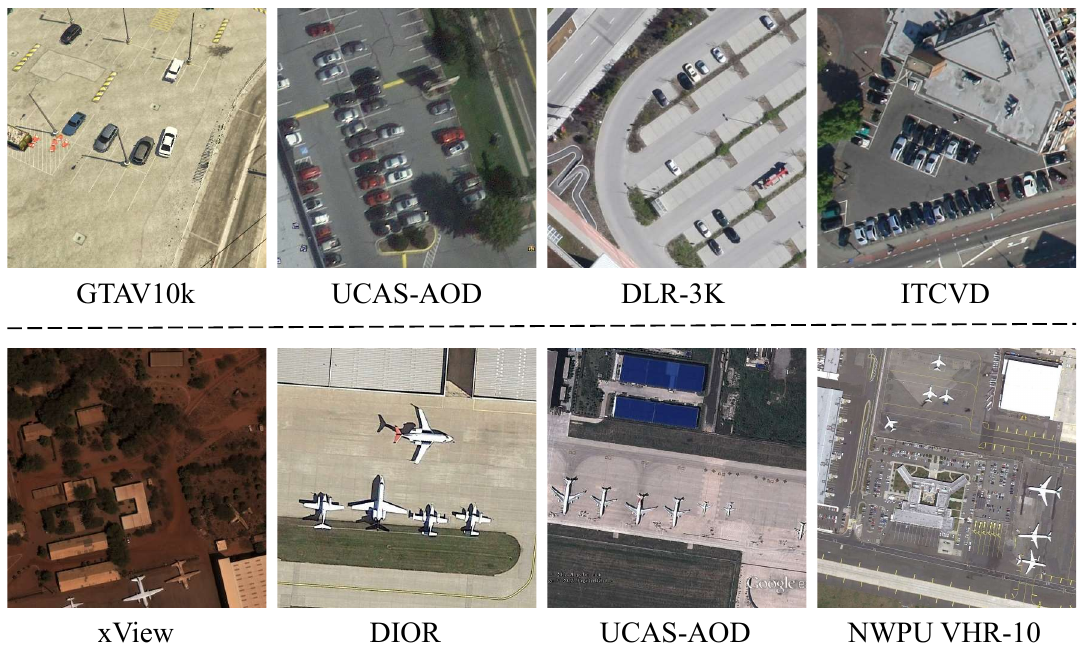}
    \caption{ Examples of synthetic-to-real adaptation (top) and cross-sensor adaptation (bottom).}
    \label{fig:0}
\end{figure} 

\begin{figure}[!t]
    \centering
    \includegraphics[width=0.95\linewidth]{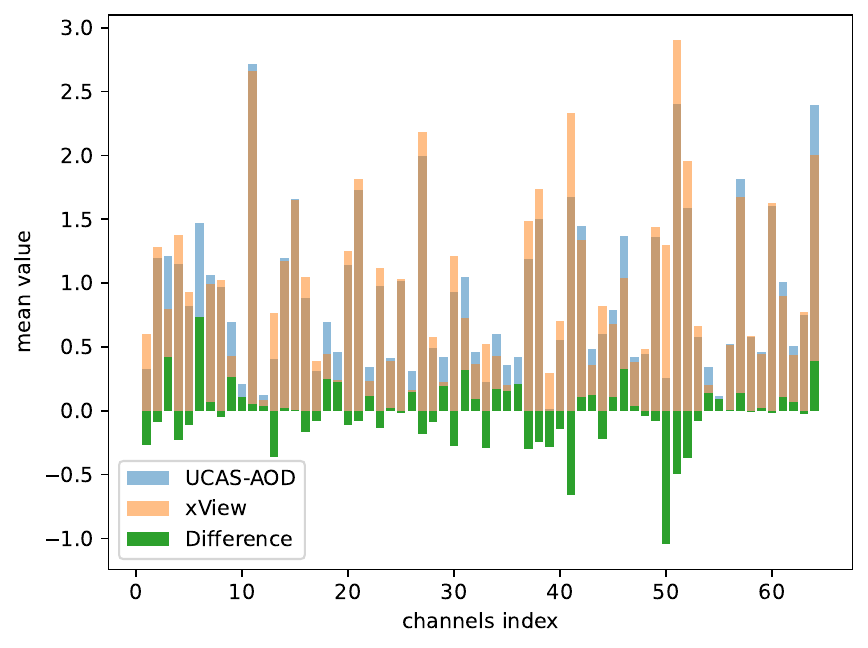}
    \caption{ Domain differences in channel statistics of low-level features. For the source domain UCAS-AOD and the target domain xView, we calculate the channel average at the first stage of the backbone. We show the first 64 (out of 256) channels for better visualization.}
    \label{fig:1}
\end{figure}

Since deep learning-based detectors are usually based on the independent and identically distributed (\textit{i.i.d}) assumption,  the detector performance drops significantly \cite{xu2022fada, zhu2023dualda} when there are domain shifts between test data and training data. Typical domain differences are intuitively reflected at the image and feature levels. Fig. \ref{fig:0} shows the domain gap in cross-sensor adaptation and synthetic\cite{liu2023unsupervised}-to-real adaptation, including image noise, image color and style, target appearance, etc. To compare the feature distribution difference in cross-sensor adaptation, we trained a Faster R-CNN model using the airplane class of UCAS-AOD, and then test the detector on xView dataset. It is generally believed that the channel statistics of low-level features can reflect the image style \cite{huang2017arbitrary, nuriel2021permuted}. As shown in Fig. \ref{fig:1}, in the shallow layer of the backbone, there are large style gaps between the source domain UCAS-AOD and the target domain xView. These domain shifts limit the application of these trained models. Therefore, discovering useful knowledge for the target domain from cheap synthetic data or existing annotated data is an urgent issue.

Recently, several unsupervised domain adaptive object detection (UDAOD) methods \cite{koga2020method,xu2022fada,liu2023unsupervised,zhu2023dualda,he2023cross}  have gradually developed in the RS field. The UDAOD aims to align the distributions of the source and target domains so that the detector trained on the supervised source domain can achieve good generalization performance on the unlabeled target domain. A classic setting of UDAOD is that both the source domain and the target domain data can be obtained during the training of domain adaptation. For example, Zhu et al. \cite{zhu2023dualda} use adversarial domain discriminators at the shallow feature and instance feature levels to determine which domain the feature comes from, thereby achieving feature distribution alignment between domains. Shi et al. \cite{shi2022unsupervised} translate optical images to SAR images using generative adversarial networks (GANs) to reduce pixel-level domain gaps. However, due to the privacy of source domain data and the high cost of data transmission, this condition is difficult to meet \cite{liang2020we}. Compared with source domain data, source domain model files are smaller and easier to transfer \cite{xia2021adaptive}. Therefore, it is more common to only be able to obtain pre-trained source domain models. How to utilize only trained source domain models and unlabeled target domain data is becoming a direction worth exploring \cite{liang2020we, xia2021adaptive}.

     \begin{figure}[tbp]
    \centering
    \includegraphics[scale=0.5]{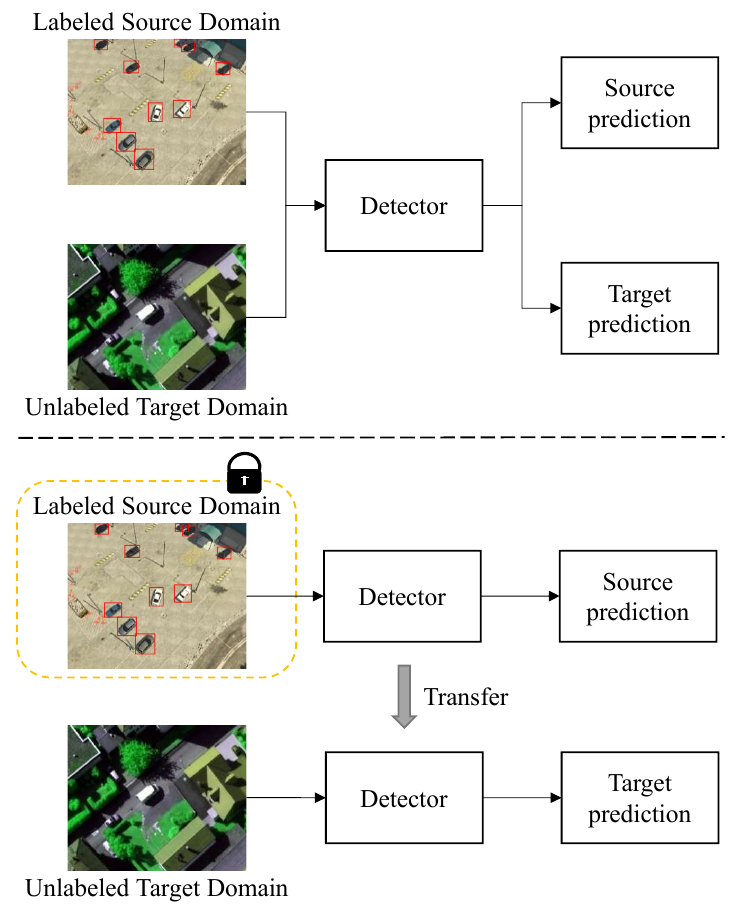}
    \caption{Adaptation settings for UDAOD (top) and SFOD (bottom). During adaptation on the target domain, the SFOD setting uses a source pre-trained detector but does not have access to the data in the source domain.}
    \label{fig:2}
    \end{figure}
    
To address this problem, researchers have proposed source-free domain adaptive object detection (SFOD) in computer vision fields \cite{li2021free,xiong2022source,chen2023exploiting, vs2023towards,vibashan2023instance}. Fig. \ref{fig:2} illustrates the difference between the SFOD and conventional UDAOD setting. Since the source domain data cannot be accessed during the transfer learning process, the widely used style transfer \cite{zhu2017unpaired} and feature alignment \cite{saito2019strong} methods in UDAOD cannot be directly applied to SFOD. Current techniques mostly rely on consistency or pseudo-labeling \cite{roychowdhury2019automatic,li2021free}. To reduce the noise of pseudo labels, some work \cite{li2022source,vs2023towards} introduced the mean teacher \cite{tarvainen2017mean} method from semi-supervised learning.  However, the existing SFOD methods are mainly designed for natural image scenarios, such as from real images to clipart and watercolor images. Domain shifts mainly occur in the high-level textures and shapes of objects. In contrast, remotely sensed images are more affected by imaging conditions, and images acquired by different sensors differ significantly in terms of imaging noise \cite{rasti2021image}, image color, and shallow style.  To make the detector more robust to domain-variant features such as color and style, several methods attempt to perturb the image with a mean image\cite{xiong2021source} or style enhancement network\cite{li2022source}. However, using an individually styled enhancement network adds additional computational burden for the training. Since it is not an end-to-end setup, gradients cannot flow to the detector.

In this paper, to cope with the problem of the unavailability of source domain data in the cross-domain object detection adaptation process, we first propose the source-free domain adaptive object detection (SFOD) setting for RS images. In the SFOD setting, we aim to achieve domain adaptation by only leveraging unlabeled target domain data and the source domain pre-trained detector. Inspired by \cite{xiong2021source}, we propose a new SFOD method based on perturbation and consistency. We believe that the characteristics of each domain include domain-invariant features and domain-varying features. Affected by imaging conditions and sensors, RS images in different domains have domain-variant features such as color and style, but domain-invariant features are shared. Therefore, domain adaptation can be achieved by focusing on domain-invariant features. We perturb domain-variant features of the target domain to construct the perturbed domain. We assume that the domain-invariant features of the perturbed domain and the target domain are estimates of the domain-invariant features of the source and target domains. By aligning the detector's behavior on the perturbed domain and the target domain, the detector is robust to domain-variant characteristics.

Our approach focuses on building the perturbed domain and aligning the detector behavior. To efficiently generate the perturbed domain, we propose the Mixed Sample Perturbation (MSP) module, which simulates the color and noise of remotely sensed image domain shifts at the pixel level. At the feature level, we propose an adversarial feature style perturbation (AFSP) module to change the feature style by changing the feature channel statistics. The AFSP module can automatically learn the style perturbation strategy and be easily inserted into the backbone network. To align the detector behavior, we introduce the Mean-Teacher \cite{tarvainen2017mean} framework, in which the teacher network with strong predictive capabilities predicts the target domain and generates pseudo labels to supervise the student network's detection of the perturbed domain. Due to the large amount of easily confusing background in RS images, pseudo labels contain many wrong predictions. To mitigate the noise of pseudo labels, we propose a prototype-based feature distillation (PFD) module.

The contributions of this article are as follows:
\begin{itemize}
	\item 
	 We introduce the source-free object detection (SFOD) setting for RS image object detection. Compared with the traditional unsupervised domain adaptive object detection (UDAOD), the SFOD setting is more practical when source domain data are unavailable.
	
	\item 
        We propose a new SFOD method, which consists of multilevel domain perturbation and multilevel alignment in the teacher-student network.

        \item 
        To efficiently generate the perturbed domain, a novel style perturbation module is designed, which can automatically learn strategies to perturb feature styles in an adversarial manner.
	
	\item 
	Experiments on synthetic-to-real and cross-sensor adaptation demonstrate the effectiveness of the proposed SFOD method on RS images. Experiments on the Sim10K\cite{johnson2016driving}-to-Cityscapes\cite{cordts2016cityscapes} adaptation demonstrate that the proposed method is also competitive in natural scenes.
\end{itemize}

\section{Related work}

\subsection{Cross-Domain Object Detection in Remote Sensing Field}
The main methods of unsupervised domain adaptive object detection in RS images can be divided into three categories, namely \textit{image-to-image translation} \cite{li2020evaluating, liu2021synthetic, shi2022unsupervised}, \textit{adversarial feature alignment} \cite{koga2020method, xu2022fada, shi2022unsupervised, liu2023unsupervised, zhu2023dualda, zou2023cross, he2023cross} and \textit{pseudo-label self-training} \cite{shi2022unsupervised, zhu2023dualda, zou2023cross, zheng2023dual}.  \textit{Image-to-image translation} is an intuitive way to explicitly transfer the style of source domain images. CycleGAN \cite{zhu2017unpaired} is a well-known work that carries out image translation without image pairing. Several CycleGAN-based methods \cite{li2020evaluating, liu2021synthetic, shi2022unsupervised, kim2022gan} have demonstrated that this kind of method can significantly reduce image style differences between domains. \textit{ Adversarial feature alignment} aligns features through the paradigm of discriminator and generator. The training process is usually implemented using gradient reversal layers (GRL) \cite{ganin2016domain}. Since domain discrepancies vary across network layers, the aligned feature levels need to change according to the adaptation scenario, including low-level features \cite{liu2023unsupervised,zhu2023dualda}, high-level features \cite{xu2022fada, shi2022unsupervised,zou2023cross} and prototype features \cite{xu2022fada, zou2023cross}. \textit{Pseudo-label self-training} methods assume that inference results with high confidence in the target domain are correct predictions. Therefore, the labels of these results can be sharpened as the ground truth for training. Many methods try to combine these three categories of methods to obtain good results, such as optical-to-SAR adaptation\cite{shi2022unsupervised, zou2023cross},  cross-time adaptation \cite{zhu2023dualda}, synthetic-to-real adaptation \cite{xu2022fada,liu2023unsupervised,zhu2023dualda}. However, these methods usually assume that source domain data is available, which limits the application scenarios and convenience of UDA object detection methods.

\subsection{Source-free Domain Adaptive Object Detection}
 As an emerging branch of domain adaptation, source-free domain adaptive object detection (SFOD) has drawn the interest of researchers in the field of computer vision \cite{li2021free,xiong2022source,chen2023exploiting, vs2023towards,vibashan2023instance}. As mentioned above, pseudo-label self-training does not depend on source domain data. This simple yet effective technique has become a classic  way of SFOD \cite{roychowdhury2019automatic,li2021free,xiong2022source}. The challenge with this kind of approach is avoiding the impact of noisy labeling. Li et al.\cite{li2021free} investigated the connection between self-entropy and the quality of pseudo-labels and suggested an automated technique for determining the confidence threshold. Xiong et al.\cite{xiong2022source} have proposed to fuse the prediction results of the prototype classifier and the original classifier as the final prediction. 
 
 Inspired by Mean-Teacher (MT) \cite{tarvainen2017mean} in semi-supervised learning, recent works \cite{chen2023exploiting,vs2023towards,vibashan2023instance} employ the exponential moving average (EMA) model as the teacher network to produce high-quality pseudo-labels on the target domain. The improvements of these methods can be summarized into three key aspects:  the type of distilled knowledge  \cite{li2022source, vs2023towards,vibashan2023instance}, the update strategy of the teacher-student network \cite{vs2022mixture, liu2023periodically}, and the target domain data perturbation \cite{xiong2021source, li2022source,yuan2022simulation}. Similar to UDAOD, Li \cite{li2022source} uses adversarial feature alignment at the image level and instance level to constrain the teacher-student network. IRG \cite{vibashan2023instance} aims to distill relationships between instance features using Graph Convolution Network (GCN).  MemCLR \cite{vs2023towards} uses memory modules to obtain robust feature patterns on the target domain. To suppress the noisy labels from one teacher, MoTE \cite{vs2022mixture} supervises the student with an ensemble of teachers generated by Monte-Carlo Dropout. To stabilize the training process, PETS \cite{liu2023periodically} presents the idea of combining dynamic teachers with static teachers. SOAP \cite{xiong2021source} first proposed the concept of the super-target domain in SFOD. It is assumed that the direction from the super-target domain to the target domain is an approximation of the direction from the target domain to the invariant domain. Thus, requiring consistency between the super-target domain and the target domain is beneficial to improving the performance of the source domain model on the target domain. Like SOAP, LODS \cite{li2022source} generates the super-target domain by using a style improvement network in place of SOAP's basic estimation of a domain-specific perturbation approach. Nevertheless, it is computationally costly to disturb the target domain using more networks. With minimal computation, our approach uses a style perturbation module, and the adversarial process sharpens the perturbation.

\section{Methodology}
In this section, we first introduce the general definition of the SFOD problem and our assumptions, then introduce the proposed framework consisting of three core modules, and finally introduce the algorithm flow and optimization method.

     \begin{figure*}[tbp]
    \centering
    \includegraphics[width=\textwidth]{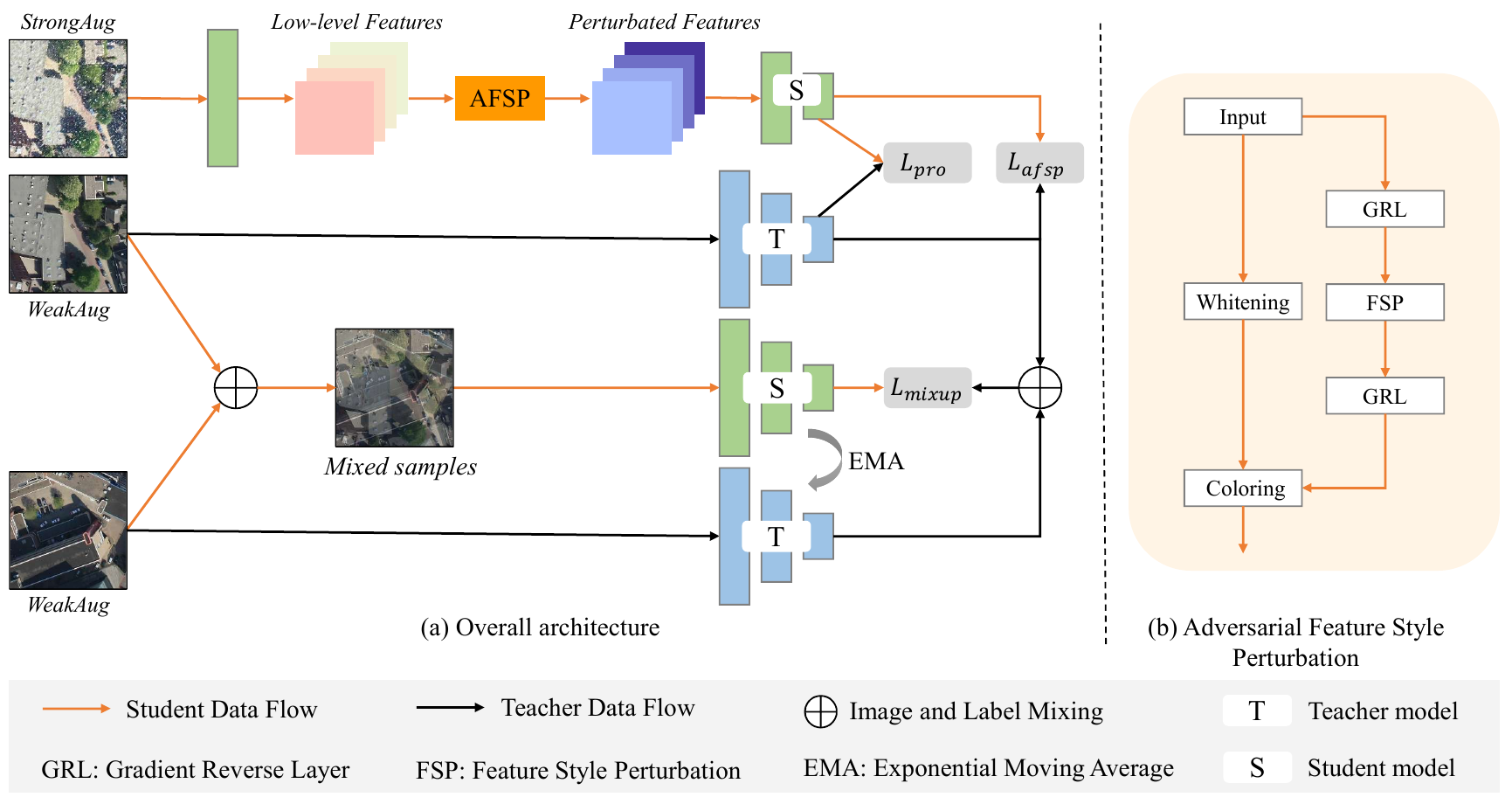}
    \caption{The framework of our proposed SFOD method. Our method can be divided into two parts: perturbed target domain generation and alignment. The image-level mixed-based sample perturbation (MSP) module and the feature-level adversarial feature style perturbation (AFSP) module aim to obtain meaningful perturbation in the target domain. To better align the behavior of detector between the target domain and the perturbed target domain, prototype-based feature distillation (PFD) is adopted at the feature level, and pseudo labeling is used at the label level. }
    \label{fig:3}
    \end{figure*}
    
\subsection{Overview}

\textbf{Problem Formulation.} We define the data in the source domain $\mathcal{D}^s$ as $\mathcal{X}^s$ , the marginal distribution as $\mathcal{P}(\mathcal{X}^s) $, and the label as $\mathcal{Y}^s$. Similarly, record the data in the target domain $\mathcal{D}^t$ as $ \mathcal{X}^t$ and the marginal distribution as $\mathcal{P}(\mathcal{X}^t) $. In the SFOD problem, the marginal distributions of the source domain and target domain are different, i.e. $\mathcal{P}(\mathcal{X}^s) \neq \mathcal{P}(\mathcal{X}^t) $. The source domain data $\mathcal{X}^s$ and labels $\mathcal{Y}^s$ are only available during pre-training of the detector model $\mathcal{M}^s$. In the domain adaptation stage, the detection model $\mathcal{M}^t$ on the target domain is  initialized using the pre-trained model $\mathcal{M}^s$.  The goal of SFOD is  to learn from the unlabeled data  of the target domain to improve the performance of the detector $\mathcal{M}^t$  on the target domain.

\textbf{Perturbed Domain Generation and Alignment.} Let the feature extractor be $\mathcal{F}$. Given the target domain image $x \in \mathcal{X}^t$, the target domain features  $\mathcal{F}(x) = \{f^t_{di},f^t_{dv}\}$ consists of domain-invariant features $f^t_{di}$ and domain-variant features $f^t_{dv}$. We perturb domain-variant features $f^t_{dv}$ while keeping domain-invariant features $f^t_{di}$ the same to generate perturbation domain $\mathcal{D}^p$. The features $\mathcal{F}_p$ of $\mathcal{D}^p$ can be formulated as
 \begin{equation}
 \mathcal{F}_p = \{f^p_{di},f^p_{dv}\} = \{f^t_{di},f^p_{dv}\}
  \end{equation}
where $f^p_{dv}$ denotes the perturbed domain-variant features. 
We then ask the detector to make consistent predictions over $\mathcal{D}^t$ and $\mathcal{D}^p$. The consistency loss can be formulated as:
 \begin{equation}
      \mathcal{L} = Dis(\mathcal{C}(\mathcal{F}(\mathcal{X}^t)),\mathcal{C}(\mathcal{F}_p(\mathcal{X}^t)))
      \label{cons}
 \end{equation}
where $\mathcal{C}$ denotes the classifier and the $Dis$ is the consistency criterion. In this way, the pre-trained detector are forced to focus on target domain-invariant features $f^t_{di}$. 
We assume that the domain-invariant features $f^{st}_{di}$ of the source and perturbed domains can be estimated from the domain-invariant features $f^{tp}_{di}$ of the target and perturbed domains.
\begin{equation}
f^{st}_{di} \leftarrow f^{tp}_{di}
\end{equation}
thereby achieving the goal of performing well on the target domain. 

\textbf{Framework.}
The proposed SFOD framework is shown in Fig. \ref{fig:3}. It consists of two domain perturbation modules: The mixed sample perturbation (MSP) module and the adversarial feature style perturbation (AFSP) module. The MSP module randomly mixes two target domain images to achieve pixel-level perturbation. The AFSP module first extracts the channel mean and standard deviation of the target domain features and predicts the adversarial style. Then, the AFSP module mixes the adversarial style and the input style as the perturbed style.

By constraining the consistency of the detector in the perturbed target domain and the target domain $\mathcal{D}^t$, the detector is forced to be insensitive to domain-variant features $f_{dv}$. The target domain image and the perturbed image are fed to the teacher and student networks, respectively. The teacher model is an ensemble of the student model with more accurate predictions. Therefore, the prediction of the target domain image (without perturbation) in the teacher model can be regarded as the supervision of the student model. In addition, the prototype-based feature distillation (PFD) module is used to align the global prototypes of the teacher and student models to mitigate the effect of noise in the pseudo-labeling. Using the pseudo-label and Mean-Teacher framework, the loss function in Eq. (\ref{cons}) can be rewritten as follows:
 \begin{equation}
 \begin{gathered}
       y_{pl} = PL(\mathcal{C}_{tea}(\mathcal{F}_{tea}(x^t))),\\
       \omega(x^t) = 1, if \; max(p(y|x^t))\geq\tau,\\
      \mathcal{L} = \sum_{x^t}\omega(x^t) \cdot {d}_{pl}(y_{pl},\mathcal{C}_{stu}(\mathcal{F}_{stu}(x^t_p)))
 \end{gathered}
  \end{equation}
 where $\omega(x^t)$ denotes the weight of sample $x^t$, $\tau$ denotes the threshold, $PL$ denotes the pseudo-labeling process in object detection, and $d_{pl}$ denotes the divergence.

\subsection{Mixed Sample Perturbation}
 Common perturbations at the image level are image augmentations, commonly used methods in the MT framework include image flipping, random image contrast, etc. However, these methods are not sufficient perturbation as they do not address the domain characteristics of the target domain data. Therefore, we attempt to use the target domain data itself to generate domain-specific perturbed samples. Specifically, considering that the pixel level domain differences in RS images mainly occur in image color and imaging noise, we use the mixed sample perturbation method to perturb the domain-variant attributes. This simple and effective method provides input gradients and Hessian regularization to encourage the model to perform linearly between training samples \cite{park2022unified}.

The original Mixup \cite{zhang2017mixup} method is mainly used in the classification task. Perturbed samples are constructed through convex combinations of training samples. The calculation formula is as follows:
\begin{equation}
\begin{aligned}
      & x_{mix} = \lambda x_i + (1 - \lambda)x_j \\
      & y_{mix} = \lambda y_i + (1 - \lambda)y_j
\end{aligned}
\label{mixup}
\end{equation}
where $x_{mix}$ denotes the mixed image, $y_{mix}$ denotes the mixed label. We set the mixing parameter $\lambda$ to 0.5 in all experiments. However, in the SFOD task, we do not have ground-truth labels for the target domain. Therefore, we use the pseudo-labels generated by the teacher detector. Let $B$ be predict bounding boxes and $C$ be hard labels. We generate the pseudo-labels $\{B, C\}_i^{pseudo}$ according to the confidence threshold $\tau$. Using the union of predicted bounding boxes in the two images as the mixed label, the formula is as follows:
\begin{equation}
      y_{mix} = \{B, C\}_i^{pseudo} \cup \{B, C\}_j^{pseudo}.
      \label{mix_label}
\end{equation}

The loss of mixed sample perturbation for object detection can be written as:
\begin{equation}
    \mathcal{L}_{mixup}=\sum_n \mathcal{L}_{cls}(x^n_{mix},y^n_{mix}) + \mathcal{L}_{reg}(x^n_{mix},y^n_{mix})
    \label{mix_loss}
\end{equation}
where $x^n_{mix}$ denotes the $n$-th mixed image.

We use a combination of basic image augmentations and the MSP module in practice to improve sample diversity. It is worth noting that in semi-supervised learning, there is a classic method of perturbing the input called virtual adversarial training (VAT) \cite{miyato2018virtual}. This method produces unrealistic samples. We also compare this adversarial perturbation method with the proposed MSP module in experiments.

\subsection{Adversarial Feature Style Perturbation}
\begin{figure}[tbp]
    \centering
    \includegraphics[scale=0.6]{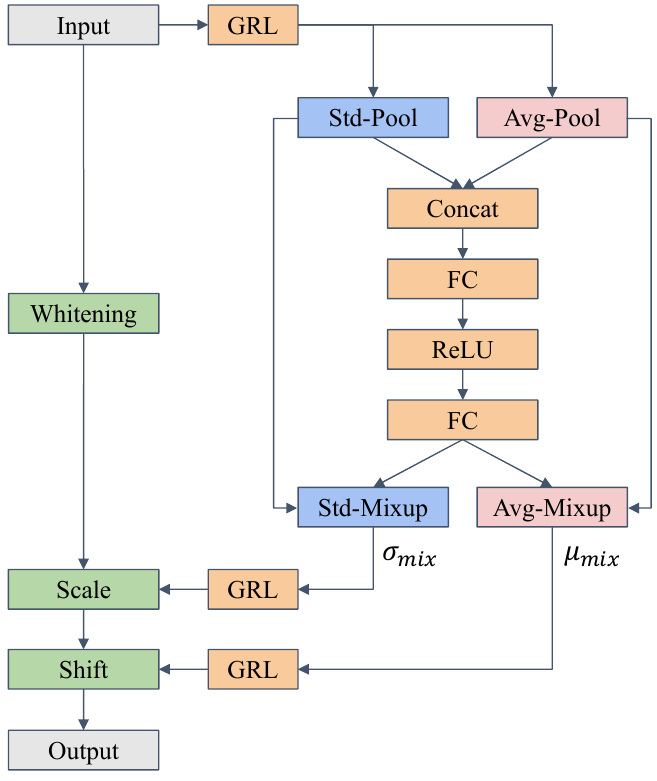}
    \caption{The schema of AFSP module.}
    \label{fig:4}
    \end{figure}

To further perturb the style of the target domain image, we also implement perturbations at the feature level. In the style transfer field, common methods to change image style include the use of the gram matrix, WCT \cite{li2017universal}, GANs, and so on. However, these methods are computationally expensive, and we would like to use simpler methods. AdaIN \cite{huang2017arbitrary} pointed out that channel statistics in convolutional neural networks (CNNs) can represent feature styles. Inspired by this, we try to perturb the channel statistics of features. To learn the perturbation strategy automatically, we propose the learnable Adversarial Feature Style Perturbation (AFSP) module. The AFSP module uses the styles of the input features to predict the adversarial styles. We equip this module with dual gradient reverse layers (GRLs). This helps to embed the original backbone. In our experiment, we use Resnet-101 \cite{he2016deep} or VGG16 \cite{simonyan2014very} as the backbone. As shown in the Fig. \ref{fig:4}, the AFSP module consists of only a few layers. Hence, the computational cost is negligible.

The original AdaIN method is applied to the style transfer task, which changes the style of the generated image by changing the feature channel-wise statistics (mean, standard deviation). The formula is as follows:
\begin{equation}
AdaIN(f_s,f_t) = {\sigma}_t \cdot \frac{f_s - {\mu}_s}{{\sigma}_s} + {\mu}_t
\end{equation}
where $f_s, f_t \in \mathbb{R}^{N\times d \times H \times W}$ denote the source feature and the target feature, and $d$ is channel numbers. However, in the SFOD setting, we cannot access the source domain data. Therefore, we design the learnable AFSP module to predict adversarial channel statistics. 

Given the input image $x$, the first stages of the backbone network extract the low-level feature $\mathcal{F}(x) \in \mathbb{R}^{N\times d \times H \times W}$, where $d$ is the number of feature channels. AFSP module first concatenates the channel mean $\mu \in \mathbb{R}^{N\times d}$ and standard deviation $\sigma \in \mathbb{R}^{N\times d}$ statistics of $\mathcal{F}(x)$, and then predicts adversarial channel statistics, which can be written as follows:
\begin{equation}
({\mu}_{adv},{\sigma}_{adv}) = FC(ReLU(FC(concat(\mu,\sigma))))
\end{equation}
where the $FC$ denotes the full connect layer.
The mixup of adversarial statistics and original statistics is used as the final statistic, calculated as follows:
\begin{equation}
\begin{aligned}
      & {\mu}_{mix} = \alpha {\mu}_{adv} + (1 - \alpha)\mu \\
      & {\sigma}_{mix} = \alpha {\sigma}_{adv} + (1 - \alpha)\sigma.
\end{aligned}
\end{equation}
Based on these channel statistics, we can obtain the style perturbed features as follows:
\begin{equation}
    {\mathcal{F}_{p}(x)} = {\sigma}_{mix} \cdot \frac{\mathcal{F}(x) - \mu}{\sigma} + {\mu}_{mix}.
\end{equation}
To limit the magnitude of the perturbation, we keep the L1-norm of the perturbed features the same as that of the input features. The output of the AFSP module can be expressed as:
\begin{equation}
    {\mathcal{F}_{afsp}(x)} = \frac{\mathcal{F}_{p}(x)}{\Vert \mathcal{F}_{p}(x) \Vert_1} \cdot \Vert \mathcal{F}(x) \Vert_1 .
    \label{afsp_feat}
\end{equation}
It is worth noting that to make the parameters of AFSP capable of adversarial learning, we use the GRL method in training. The GRL inserted before the AFSP module is to avoid passing the adversarial gradient to the previous layers.

Let the weak augmentation of the image $x$ be $x_w$ and the strong augmentation of the image be $x_s$, which are input into the teacher and student model respectively. Similar to the MSP module above, we use the pseudo-labels of the teacher network $y_w$ to supervise the style-perturbed samples. The loss of adversarial feature style perturbation for object detection can be written as:
\begin{equation}
    \mathcal{L}_{afsp}=\sum_n \mathcal{L}_{cls}(x^n_{afsp},y^n_{w}) + \mathcal{L}_{reg}(x^n_{afsp},y^n_{w})
    \label{loss_afsp}
\end{equation}
where $x^n_{afsp}$ denotes the $n$-th style perturbed image.

\subsection{Prototype-based Feature Distillation} 
In existing SFOD methods, pseudo labels and consistency are often used to supervise student networks. However, due to the complex background of RS images, there is a large amount of background in the image that is easily confused with the foreground. These methods will be affected by incorrect predictions, especially when the source domain pre-trained model performs poorly in the target domain. We aim to extract more robust knowledge from the teacher network for stable training. Inspired by the Hint learning \cite{chen2017learning} in semi-supervised learning, we proposed to distill prototype-based knowledge at the feature level, which also helps bring benefits through deep supervision. We first average the features of each category in the teacher and student detectors to obtain the category prototypes, and then consider the prototypes in the historical iterations to obtain the current global prototype. Then, the prototype features of the teacher-student network are required to be consistent. In practice, we combine the prototype feature alignment and pseudo-labeling techniques to achieve multi-level supervision. 

Formally, given the backbone feature $F^s$ and $F^t$ of the student and the teacher model, the corresponding proposal boxes and scores $\{B^s, C^s\}$,  and $\{B^t, C^t\}$. To make the feature values in the student model match to the teacher's, we add a transformation layer $TF$ (a $3 \times 3$ convolution) in the student model. The mean feature of the $c$-th class in the student and the teacher model can be presented as follows:
\begin{equation}
\begin{aligned}
    &f^s_{c,i} = \frac{1}{HW}\sum_{h,w}RoIAlign(TF(F^s),B^s_{c,i}), \\
    &f^t_{c,j} = \frac{1}{HW}\sum_{h,w}RoIAlign(F^t,B^t_{c,j}) 
\end{aligned}
\end{equation}
where $RoIAlign$ denotes the RoI Align layer. The $c$-th class local prototype ${LP}^c \in \mathbb{R}^{1 \times d}$ is calculated as follows:
\begin{equation}
    {LP}_c = \frac{1}{n_c}\sum_{i=1}^{n_c} f^s_i
\end{equation}
where $n_c$ denotes the numbers of boxes in the $c$-th class. The global prototype at $i$-th iteration is updated as follows:
\begin{equation}
    {GP}_{c,i} = \beta{LP}_{c,i} + (1-\beta){LP}_{c,{i-1}}.
    \label{gp}
\end{equation}
We set the update parameter $\beta$ to be 0.7 in all experiments. The loss of prototype-based feature distillation can be written as:
\begin{equation}
    \mathcal{L}_{pro}=\sum_c {\Vert {GP}^t_c - {GP}^s_c \Vert_2}.
    \label{loss_pro}
\end{equation}

    \begin{algorithm}[tbp]
	\caption{Perturbed Domain Generation and Alignment}\label{alg:alg1}
	
	\begin{algorithmic}[1]
		\STATE \textbf{Input:} target domain $\mathcal{D}^t$, pre-trained detector $\mathcal{M}^s$ based on source domain $\mathcal{D}^s$, threshold $\tau$ of pseudo-labeling, batch size $N$, EMA period $ema\_period$. 
		\STATE \textbf{Initialize:} Use $\mathcal{M}^s$ to initialize the parameters of student  $\mathcal{S}$ and  teacher $\mathcal{T}$. Initialize AFSP module, the transformation layer $TF$ of PFD module, and global prototypes ${GP}_s,{GP}_t$.
		
		\FOR{$iter \gets 0$ to $max\_iter$ }
	    \STATE $x_1 \leftarrow Sample(N/2, \mathcal{D}^t$), $x_2 \leftarrow Sample(N/2, \mathcal{D}^t$)
            \STATE Generate pseudo-labels with $\tau$:  \\
            $Y_1 \leftarrow  \mathcal{T}(x_1), Y_2 \leftarrow  \mathcal{T}(x_2)$ 
            \STATE \textbf{Pixel-level pertubation}: \\
		 Generate mixed samples by Eq. (\ref{mixup}), \\
             Calculate mixed labels by Eq. (\ref{mix_label}), \\
		 Calculate $\mathcal{L}_{mixup}$ by Eq. (\ref{mix_loss});
            \STATE \textbf{Feature-level pertubation}:\\
             Generate $\mathcal{F}_{afsp}$ by Eq. (\ref{afsp_feat}),\\
	     Calculate $\mathcal{L}_{afsp}$ by Eq. (\ref{loss_afsp});
            \STATE \textbf{Feature-level distillation}:\\
             Update ${GP}_s$ and ${GP}_t$ by Eq. (\ref{gp}),\\
             Calculate $\mathcal{L}_{pro}$ by Eq. (\ref{loss_pro})
            \STATE Optimize student $\mathcal{S}$ by Eq. (\ref{total_loss})
            \STATE \textbf{Adversarial learning}:\\
             Optimize the AFSP module by Eq. (\ref{adv_loss})
            \IF{$iter \bmod ema\_period = 0$ }
             \STATE Update teacher  $\mathcal{T}$ by Eq. (\ref{ema})
            \ENDIF
		\ENDFOR
		
	\end{algorithmic}
	\label{alg1}
    \end{algorithm} 

\subsection{Overall Objective}
We utilize the proposed MSP, AFSP, and PFD modules in the MT framework for SFOD on RS images. The SFOD optimization algorithm is shown in Algorithm \ref{alg1}. The overall objective can be written as:
\begin{equation}
    \mathcal{L}_{total} =  \mathcal{L}_{mix} + \mathcal{L}_{afsp} + \gamma \mathcal{L}_{pro}
    \label{total_loss}
\end{equation}
where $\gamma$ denotes the balance weight of $\mathcal{L}_{pro}$. We set $\gamma$ to 0.5 in all experiments. The adversarial style perturbation loss can be written as:
\begin{equation}
    \mathcal{L}_{adv} = -\mathcal{L}_{afsp}.
    \label{adv_loss}
\end{equation}
The parameters of the teacher model are updated by student model as follows:
\begin{equation}
\theta_{tea} = \eta \cdot \theta_{tea} + (1-\eta)\cdot\theta_{stu}
\label{ema}
\end{equation}
where $\eta$ denotes the update rate.

\section{Experiments}
In this section, we first introduce the experimental datasets and evaluation metrics, then qualitatively and quantitatively compare the performance of the proposed method with SOTA on multiple datasets.  Finally, we analyze and discuss the impact of each core module and hyperparameters through  detailed ablation experiments.

\subsection{Datasets}
 This article focuses on two tasks: synthetic-to-real vehicle adaptation and cross-sensor airplane adaptation. We used six commonly used real-world datasets (UCAS-AOD \cite{zhu2015orientation}, DLR-3K \cite{liu2015fast}, ITCVD \cite{yang2019vehicle}, DIOR \cite{li2020object}, xView \cite{lam2018xview}, NWPU VHR-10 \cite{cheng2014multi}) in RS image object detection task, a synthetic vehicle dataset GTAV10k \cite{liu2023unsupervised}. To compare with more methods in the field of computer vision, we also validate the proposed method using the synthetic car detection dataset Sim10k\cite{johnson2016driving} and the real city scene dataset Cityscapes\cite{cordts2016cityscapes}. The Table \ref{table.1} shows the dataset usage for each task.

\textit{1)} \textit{GTAV10k} $\rightarrow$ \textit{UCAS-AOD, DLR-3K, ITCVD}: The \textbf{GTAV10k} dataset contains 10001 synthetic images that mimic an aerial perspective. The image size is  $1280 \times 1024$. The \textbf{UCAS-AOD} dataset contains 510 images of the vehicle category, which are derived from Google Earth. The image sizes are mainly $1280\times659$ and $1372\times941$. The \textbf{DLR-3K} dataset contains 20 images, which are taken over Munich, Germany. Each image has $5616\times3744$ pixels with a resolution of 13cm. The \textbf{ITCVD} dataset contains 173 images with a resolution of 10cm, taken over Enschede, The Netherlands. Each image has $5616\times3744$ pixels. In our synthetic-to-real vehicle adaptation experiment, the GTAV10k dataset is used as the source domain. The DLR-3K dataset, the ITCVD dataset, and the vehicle category of UCAS-AOD datasets are used as the target domain, respectively. To facilitate the training, we crop the original large images of DLR-3K and ITCVD into $600\times 600$ patches.

\textit{2)} \textit{DIOR, UCAS-AOD, NWPU VHR-10} $\rightarrow$ \textit{xView}: \textbf{xView} is a large-scale object detection dataset covering multiple areas of the world, containing 1 million of objects in 60 categories. The image was collected from the WorldView-3 satellite with a resolution of 0.3m. Image size mainly ranges from $2500 \times 2500$ to $5000 \times 3000$. The \textbf{DIOR} dataset contains 1387 airplane category images, which are collected from Google Earth. The image size is $800 \times 800$. The \textbf{UCAS-AOD} dataset contains 1000 images of airplane category. The image sizes are mainly $1280 \times 659$ and $1372 \times 941$. The \textbf{NWPU VHR-10} dataset contains 800 images, which are collected from Google Earth and the Vaihinge dataset. Image size ranges from $543 \times 430$ to $1728 \times 1028$. In our cross-sensor airplane adaptation experiment, the airplane category of DIOR, UCAS-AOD, and NWPU VHR-10 datasets were used as source domain data to pre-train the detector. The four categories (small aircraft, helicopter, fixed-wing aircraft, and cargo plane) of XView are merged into the airplane category as the target domain. We crop the images of xView into $600\times 600$ patches.

\textit{3)} \textit{Sim10k} $\rightarrow$ \textit{Cityscapes}: \textbf{Sim10k} is a natural-view car dataset containing 10000 synthetic images. \textbf{Cityscapes} is an 8-category object detection dataset of city scenes, containing 3475 images. In the synthetic-to-real car adaptation experiment, we use Sim10k as the source domain and the vehicle category of Cityscapes as the target domain.

\begin{table*}[t]\caption{Datasets useage for evaluation of source-free domain adaptive object detection methods.}
    \label{table.1}
    \centering
    \resizebox{\linewidth}{!}{
    \begin{tabular}{cccccc}
    \toprule
   \textbf{Adaptation}  &\textbf{Task}               &\textbf{Source domain} &\textbf{Source images collection}    &\textbf{Target domain}  &\textbf{Target images collection}   \\
    \midrule
    \multirow{3}{*}{synthetic-to-real} 
      &overhead vehicle detection         &GTAV10k &GTA V engine      &UCAS-AOD  &Google Earth      \\ 
     &overhead vehicle detection         &GTAV10k  &GTA V engine    &DLR-3K    &DLR 3K camera    \\ 
      & overhead vehicle detection         &GTAV10k &GTA V engine     &ITCVD   & airplane platform    \\ 
      \midrule
    \multirow{3}{*}{cross-sensor} 
         & overhead airplane detection        &DIOR  &Google Earth       &xView   &WorldView-3 satellites     \\ 
       & overhead airplane detection        &UCAS-AOD  &Google Earth   &xView   &WorldView-3 satellites     \\ 
        &overhead airplane detection        &NWPU VHR-10 &Google Earth, Vaihinge dataset &xView   &WorldView-3 satellites     \\ 
    \midrule
    synthetic-to-real  &natural car detection             &Sim10k &GTA V engine &Cityscapes     & OnSemi AR0331  \\
    \bottomrule 
    \end{tabular}}
    \end{table*}

\subsection{Implementation Details}
 Following the settings of \cite{saito2019strong,liu2023unsupervised,vibashan2023instance}, we use Faster R-CNN \cite{ren2015faster} as the experimental framework for object detection. By default, Resnet-101 \cite{he2016deep} is used as the backbone. In Sim10k  $\rightarrow$ Cityscapes experiment, VGG16\cite{simonyan2014very} is used as the backbone. We add the AFSP module after stage $1$ of the backbone of the student network. We set the default value of $\alpha$ in the AFSP module to 0.5. We insert the AFSP module after stage 1 of the backbone. In cross-sensor adaptation experiments, $\alpha$ is set to 0.3. In both the pre-training and domain adaptation stages, we set the initial learning rate for 5 epochs to 0.001, and then decay the learning rate for 2 epochs to 0.0001. In the pre-training phase, we set the batch size to 1. In the domain adaptation experiment, the batch size is set to 2, which means that two images are fed at each iteration to generate mixed samples.

In the MT framework, weak augmentation is horizontally flipping, while strong augmentation includes AutoContrast, Brightness, Color, Contrast, GrayScale, and GaussBlur. The threshold $\tau$ in pseudo-labeling is 0.7. We set the EMA update parameter to 0.9. Update teacher network parameters once after each epoch. All short edges of input images are scaled to 600 pixels. We use mAP with an intersection over union (IoU) threshold of 0.5 as the evaluation metric. All experiments are carried out on an 11 GB NVIDIA RTX 2080ti graphics card.

\subsection{Comparisons with State-of-the-Art Methods}
To verify the effectiveness of the proposed method, we compare the methods in UDAOD and SFOD, as follows:

\textit{1) UDAOD Methods}: DA-Faster \cite{chen2018domain}, SWDA \cite{saito2019strong}, HTCN \cite{chen2020harmonizing}, SCL \cite{shen2019scl}, MeGA-CDA \cite{vs2021mega}, TDD \cite{he2022cross}, TIA \cite{zhao2022task}, and DSCR \cite{liu2023unsupervised}. Specifically, DA-Faster, SWDA, HTCN, and SCL methods mainly adversarially align source domain and target domain features. MeGA-CDA  aligns the class prototypes using the memory module. TDD preserves domain-specific characteristics through two detection branches, while DSCR captures domain-specific information with two channel recalibration branches. TIA performs alignment in task spaces using auxiliary heads.

\textit{2) SFOD Methods}: SOAP \cite{xiong2021source}, PLA-DAP \cite{xiong2022source}, MT \cite{tarvainen2017mean}, MT+VAT \cite{miyato2018virtual}, IRG \cite{vibashan2023instance}, MemCLR \cite{vs2023towards}, and LPU \cite{chen2023exploiting}. Most SFOD methods are based on MT. Specifically, SOAP incorporates domain-specific perturbations and adversarial feature discriminators. PLA-DAP uses global prototypes to implicitly align the source and target domains. IRG distills the relation of objects with instance relation graphs. MemCLR uses global memory banks and contrastive loss to learn better feature representation. LPU leverages the low-confidence pseudo-labels through soft training. 

We use MT with pseudo-labels as the baseline. MT+VAT is obtained by perturbing the image in the form of VAT\cite{miyato2018virtual} at the pixel level, as a comparison with our perturbation method. For a fair comparison, we reproduced all of the comparison methods so that the backbone networks used are the same, and other parameters are set according to the recommendations of relevant papers.

 \begin{table}[t]\caption{Vehicle detection results from GTAV10k to UCAS-AOD, DLR-3K, and ITCVD, respectively.}
    \label{table.2}
    \centering
    \resizebox{\linewidth}{!}{
    \begin{tabular}{c|c|c|c|c}
    \toprule
   Settings &Methods               &  UCAS-AOD    &DLR-3K     & ITCVD  \\
    \midrule
    Source only &Faster R-CNN           &52.8      & 57.8      &45.7   \\ \midrule
    \multirow{6}{*}{UDAOD}  
    &DA-Faster       &58.0      &56.2      &50.7   \\  
    &SWDA                  &57.7      &62.5      &51.0   \\
    &HTCN                  &54.7      &67.5      &46.6   \\
    &SCL                   &60.7      &67.5      &52.4   \\
    &TIA                   &56.6      &63.0      &48.9   \\
    &DSCR                  &66.5      &68.3      &58.0   \\ \midrule
    \multirow{6}{*}{SFOD}  
    &SOAP                  &64.9      &64.1      &57.2   \\ 
    &PLA-DAP               &58.3      &63.8      &52.4   \\ 
    &MT                    &64.4      &68.8      &58.2   \\ 
    &MT+VAT                &64.9      &69.5      &59.2 \\ 
     &IRG                  &\underline{65.4}     &\underline{71.2}  &59.8   \\ 
    &MemCLR                &65.0      &70.8      &\underline{61.9}   \\ \midrule
   SFOD &Ours           &\textbf{73.6}  &\textbf{72.4} &\textbf{72.3}   \\ 
    \bottomrule 
    \end{tabular}}
    \end{table}

    \begin{figure*}[tbp]    
      \centering        
      \subfloat[GTAV10k $\rightarrow$ UCAS-AOD]  
      {
          \label{fig:5subfig1}\includegraphics[width=0.33\textwidth]{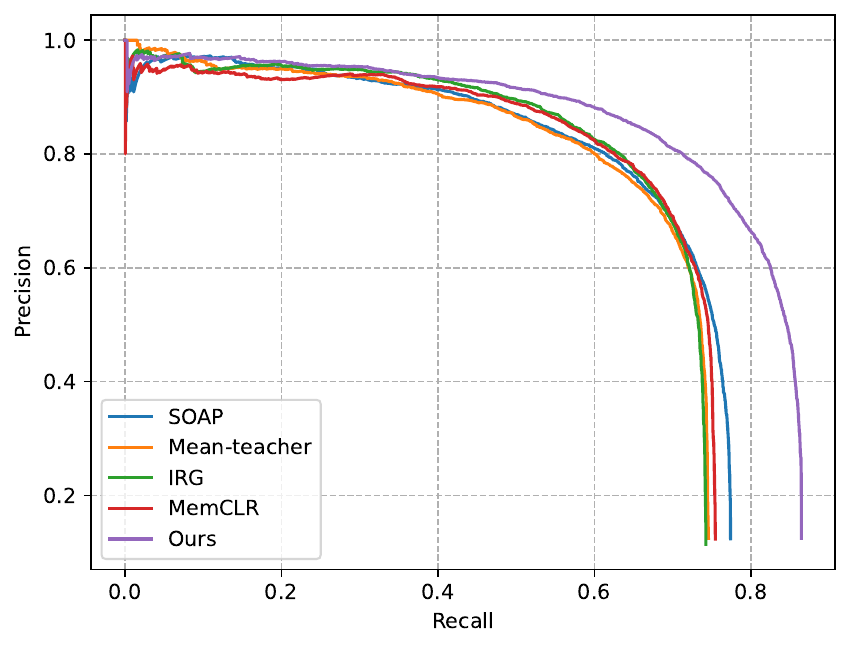}
      }
      \subfloat[GTAV10k $\rightarrow$ DLR-3K]
      {
          \label{fig:5subfig2}\includegraphics[width=0.33\textwidth]{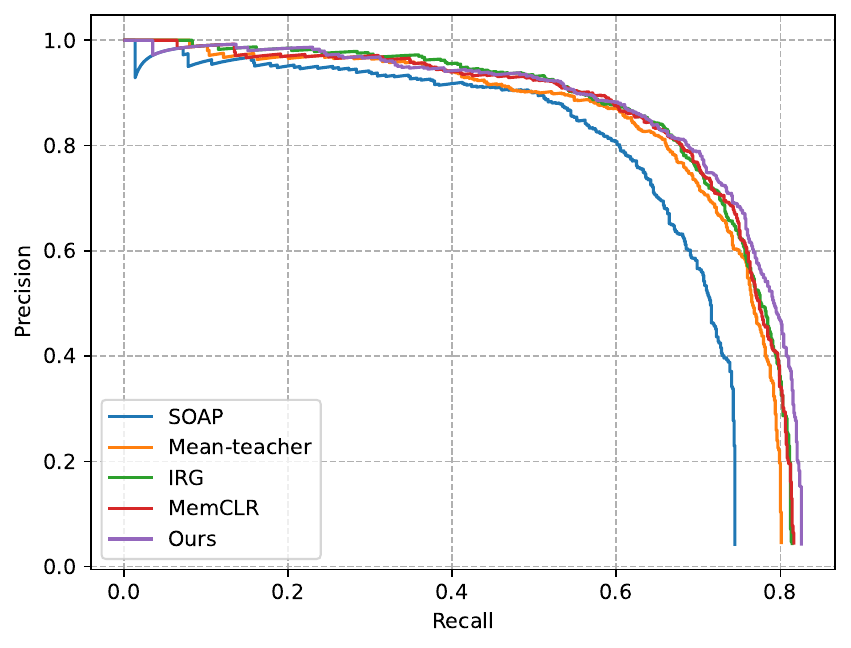}
      }
      \subfloat[GTAV10k $\rightarrow$ ITCVD]
      {
          \label{fig:5subfig3}\includegraphics[width=0.33\textwidth]{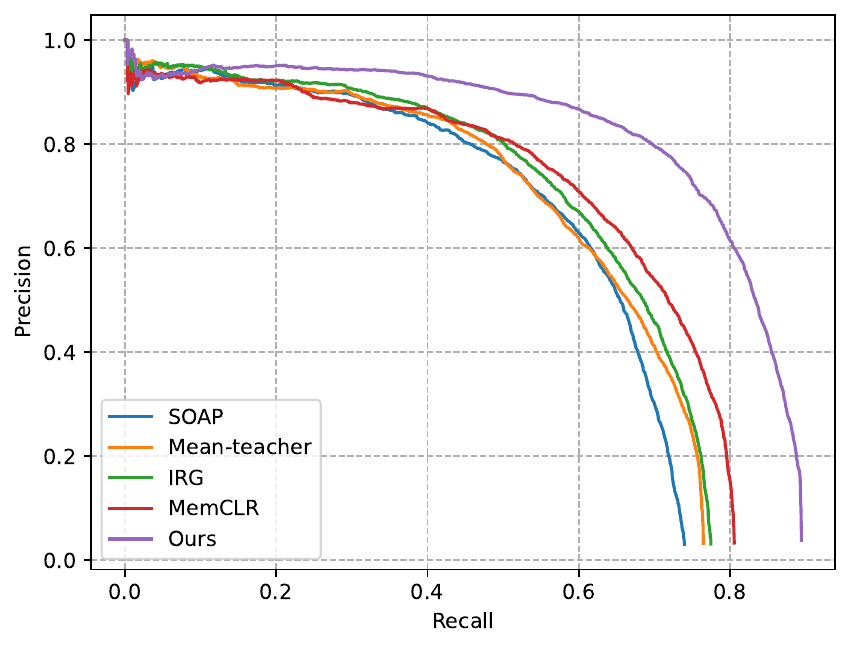}
      }
      \caption{PR curves of three synthetic-to-real adaptation experiments.}    
      \label{fig:5}       
    \end{figure*}

    \begin{figure*}[tbp]    
      \centering           
      \subfloat[Source Only]   
      {
          \label{fig:6subfig1}\includegraphics[width=0.33\textwidth]{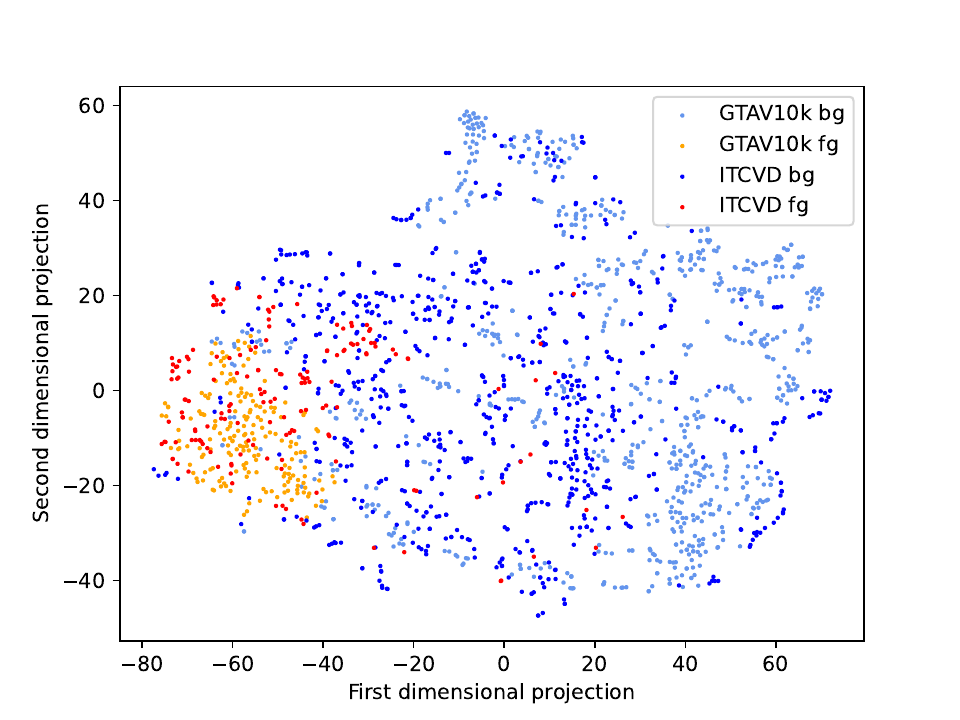}
      }
      \subfloat[MT]
      {
          \label{fig:6subfig2}\includegraphics[width=0.33\textwidth]{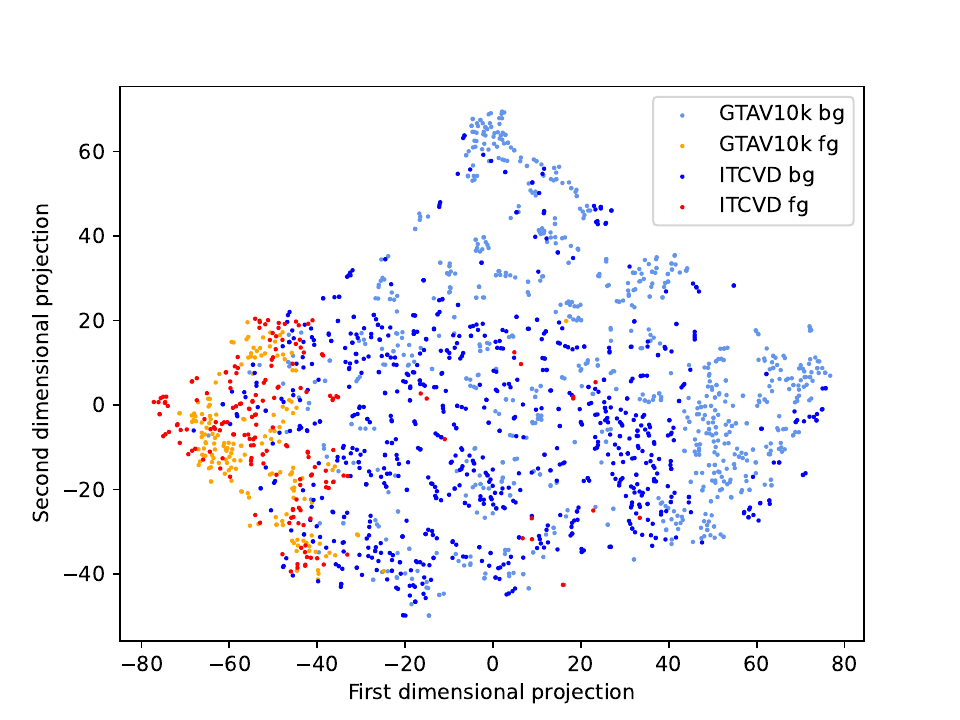}
      }
      \subfloat[Ours]
      {
          \label{fig:6subfig3}\includegraphics[width=0.33\textwidth]{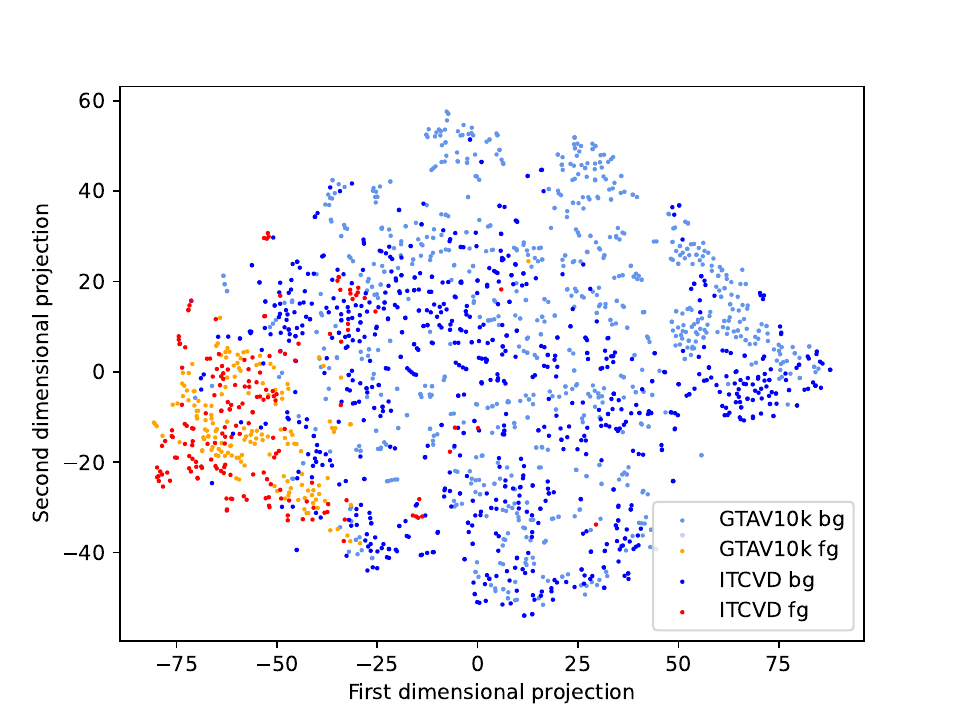}
      }
      \caption{Visualization of feature in GTAV10k $\rightarrow$ ITCVD experiment via tSNE. The bg represents the feature of backgrounds rois. The fg represents the feature of foreground rois.}   
      \label{fig:6}       
    \end{figure*}
    
   \begin{figure*}[tbp]
    \centering
    \includegraphics[width=\linewidth]{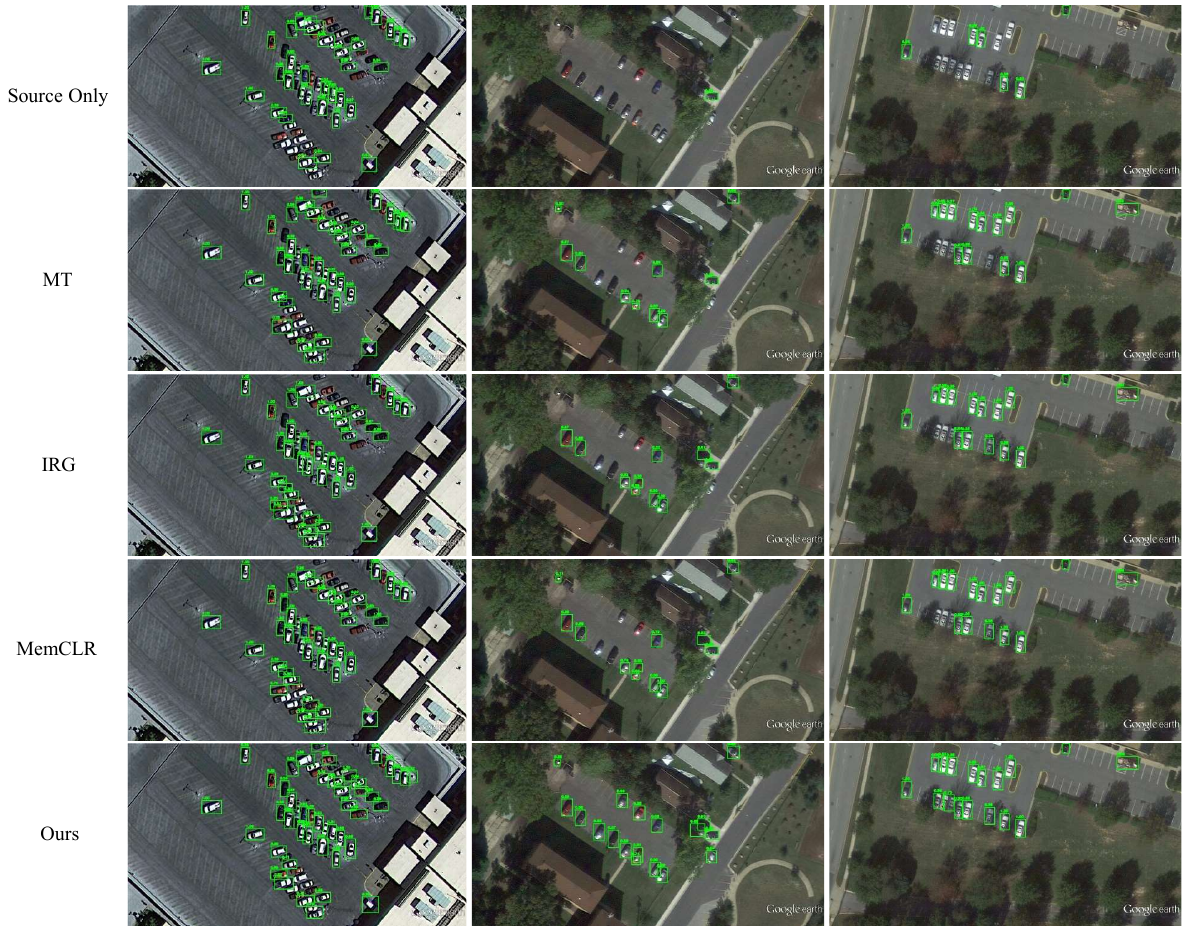}
    \caption{Qualitative comparison on GTAV10k $\rightarrow$ UCAS-AOD vehicle detection. From top to bottom: Source only, MT, IRG, MEMCLR, Ours.}
    \label{fig:7}
    \end{figure*}
    
\textit{1)} \textit{Results on GTAV10k} $\rightarrow$ \textit{UCAS-AOD, DLR-3K, ITCVD}: Table \ref{table.2} compares the results of different methods in the three synthetic-to-real adaptation experiments. Our method achieved the best results compared to other SFOD methods. Compared with the source domain pre-trained model without any adaptation, the proposed SFOD method achieved significant performance improvements (+20.8\% AP, 14.6\% AP, 26.6\% AP) by only utilizing unlabeled data on the target domain. Compared with the UDAOD method, our method achieves better results in the more difficult setting. This is very convenient for practical use where the source domain data is inconvenient to transfer and obtain. We also compared a similar adversarial perturbation method VAT. MT+VAT is only slightly improved based on MT. This indicates that using the proposed method to perturb domain-variant features is more effective in the adaptation from synthetic to real RS images.

Fig. \ref{fig:5} compares the PR curves of 5 SFOD methods in three synthetic-to-real experiments. Our method has the largest area under the PR curve. Especially in the high recall area, there is a significant improvement. As shown in Fig. \ref{fig:6}, we used the t-SNE \cite{van2008visualizing} method to analyze the feature distribution of the pre-trained model and domain adaptation model on the source domain and target domain. It can be seen that the pre-trained model completely separates foreground and background in the source domain, but incorrectly classifies a large amount of foreground as background in the target domain. There is also a significant deviation in the overall distribution of the two domains. The MT model aligns the two domain distributions as a whole, but there is a large amount of target domain foreground and background around the classification boundary. Our method effectively separates the foreground and background of the target domain, although the source domain data is not exposed in SFOD. Fig. \ref{fig:7} visualizes the vehicle detection results on the UCAS-ADO dataset. It can be seen that our method significantly improves the vehicle recall rate. Our method works well even in difficult scenes, such as dense arrays of objects (left), high reflections in parts of the vehicle (middle), and interference from twigs and shadows (right). In contrast, source domain models can only detect a small number of color types of vehicles. This indicates that the generalization performance of the detector is not satisfactory when there is a domain gap, so domain adaptation on the target domain is required. 

    \begin{table}[t]\caption{Airplane detection results from DIOR, UCAS-AOD, NWPU VHR-10 to Xview, respectively.}
    \label{table.3}
    \centering
     \resizebox{\linewidth}{!}{
    \begin{tabular}{c|c|c|c|c}
    \toprule
  Settings  &Methods               & DIOR  &UCAS-AOD     &NWPU VHR-10 \\ \midrule
    Source only  &Faster R-CNN          & 59.9      & 61.2      &47.1   \\ \midrule
    \multirow{4}{*}{UDAOD}  
    &SWDA                  & 67.7      & 68.9      &48.9   \\
    &HTCN                  & 71.7      & 69.4      &46.9   \\
    &SCL                   & 72.3      & 68.8      &48.9   \\
    &DSCR                  & 70.9      & 68.8      &51.0   \\ \midrule
    \multirow{6}{*}{SFOD}  
    &SOAP                  & \underline{68.4}      & 70.8      &52.7   \\ 
    &PLA-DAP               & 58.7      & 62.2      &46.6   \\ 
    &MT                    & 65.9      & 73.5      &55.0   \\ 
    &MT+VAT                & 66.9      & 73.2      &54.8 \\ 
    &IRG                   & 67.6      & 75.2      &57.1   \\ 
    &MemCLR                & 67.6      &\textbf{76.3} &\underline{57.6}   \\ 
    \midrule
   SFOD &Ours           &\textbf{72.6}     &\underline{76.2} &\textbf{60.7}   \\ 
    \bottomrule 
    \end{tabular}}
    \end{table}
    
  \begin{figure*}[tbp]    
  \centering            
  \subfloat[DIOR $\rightarrow$ xView]  
  {
      \label{fig:8subfig1}\includegraphics[width=0.33\textwidth]{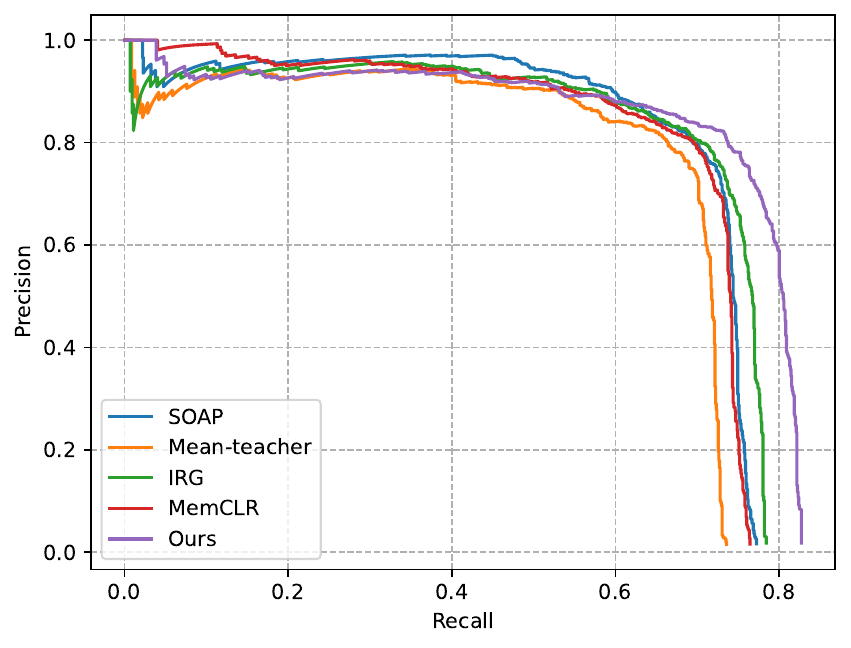}
  }
  \subfloat[UCAS-AOD $\rightarrow$ xView]
  {
      \label{fig:8subfig2}\includegraphics[width=0.33\textwidth]{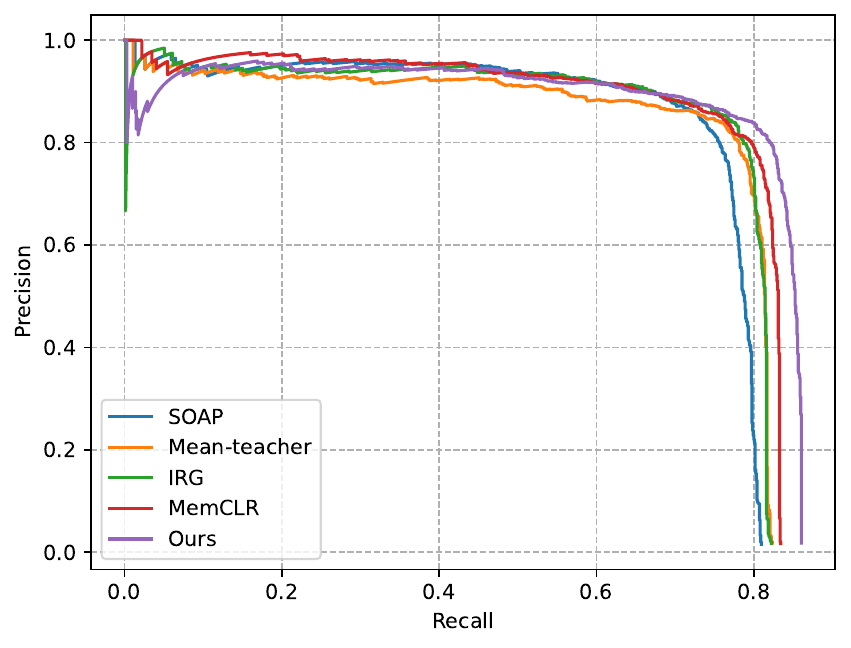}
  }
  \subfloat[NWPU VHR-10 $\rightarrow$ xView]
  {
      \label{fig:8subfig3}\includegraphics[width=0.33\textwidth]{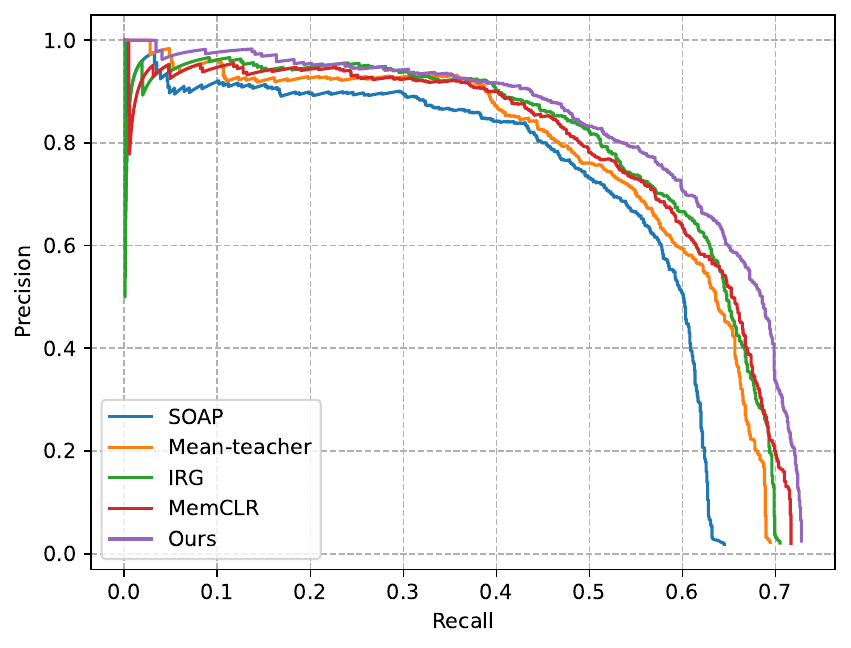}
  }
  \caption{PR curves of three cross-sensor adaptation experiments.}    
  \label{fig:8}       
\end{figure*}

     \begin{figure*}[tbp]
    \centering
    \includegraphics[width=\linewidth]{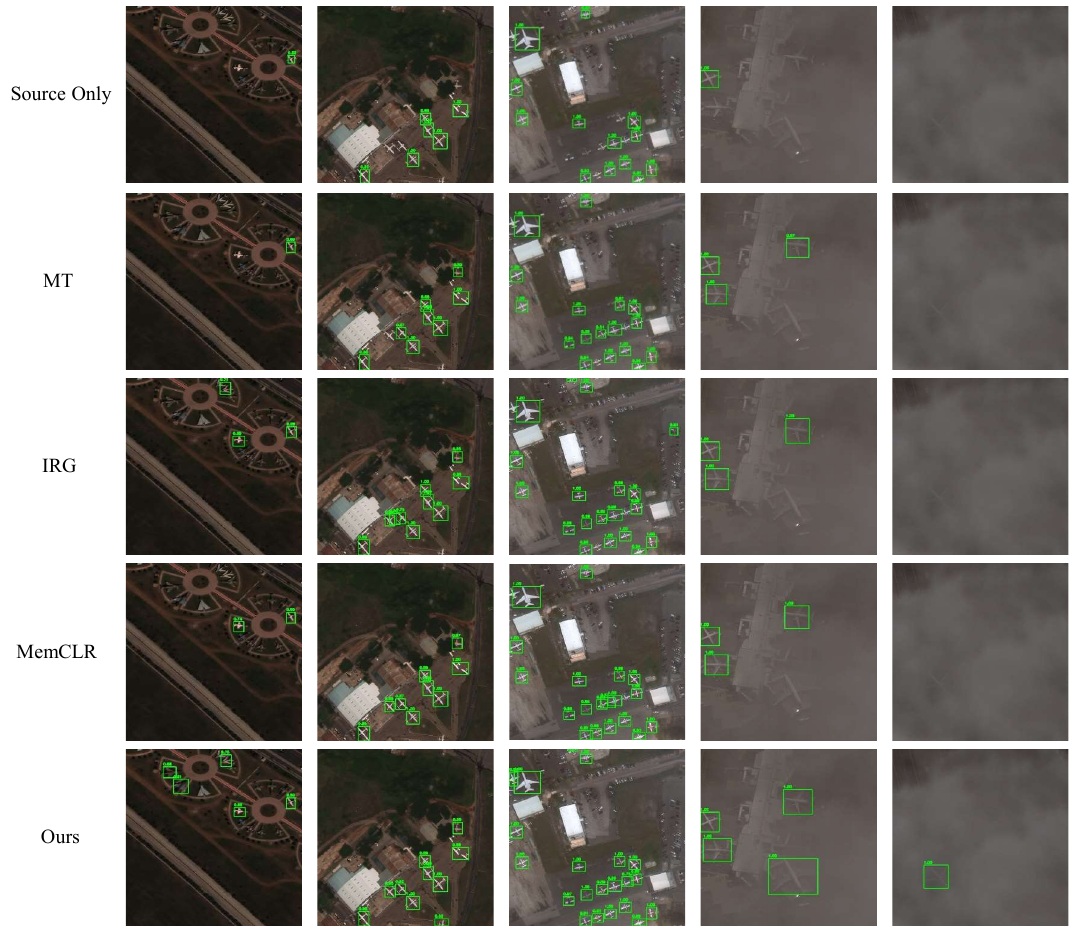}
    \caption{Qualitative comparison on DIOR $\rightarrow$ xView airplane detection. From top to bottom: Source only, MT, IRG, MEMCLR, Ours.}
    \label{fig:9}
    \end{figure*}
    
\textit{2)} \textit{Results on DIOR, UCAS-AOD, NWPU VHR-10} $\rightarrow$ \textit{xView}:
Table \ref{table.3} shows the results of three experiments of cross-sensor adaptation. Our method achieves the best on both DIOR and NWPU VHR-10, and the second highest on UCAS-AOD (76.2\%AP versus 76.3\%AP), which confirms the effectiveness of our method. By comparing the three experiments horizontally, it can be found that the model trained on NWPU VHR-10 performs the worst in the target domain xView. It indicates that there is a large difference between the two domains, while the domain gaps of DIOR and UCAS-AOD to xView are relatively small. The UDAOD methods are competitive with SFOD methods in DIOR $\rightarrow$ xView and UCAS-AOD $\rightarrow$ xView while having almost no impact on NWPU VHR-10 $\rightarrow$ xView. In contrast, our method shows significant results in all three experiments.

We also visualize the PR curves in Fig. \ref{fig:8} and aircraft detection results in Fig. \ref{fig:9} of cross-sensor adaptation. The xView dataset contains a variety of aircraft targets, including various colors, brightness, and imaging conditions. These factors seriously affect the detection performance of the source domain model. The proposed method can detect aircraft targets in thin clouds and mist.

    \begin{table}[t]\caption{Car detection results from Sim10k to Cityscapes.}
    \label{table.4}
    \centering
     \resizebox{\linewidth}{!}{
    \begin{tabular}{ccccc}
    \toprule
   Settings &Methods           &Venue             &Backbone  & mAP    \\ \midrule
    Source only &Faster R-CNN       &NeurIPS, 2015 &VGG16     &37.7    \\ \midrule
    \multirow{5}{*}{UDAOD}  
    &SWDA\cite{saito2019strong}     &CVPR, 2019 &VGG16     &40.1    \\
    &HTCN\cite{chen2020harmonizing} &CVPR, 2020 &VGG16     &42.5    \\
    &SCL\cite{shen2019scl}          &arXiv, 2019 &VGG16     &42.6    \\
    &MeGA-CDA\cite{vs2021mega}      &CVPR, 2021 &VGG16     &44.8   \\ 
    &TDD\cite{he2022cross}          &CVPR, 2022 &VGG16     &53.4    \\ \midrule
    \multirow{7}{*}{SFOD}  
    &SOAP\cite{xiong2021source}     &IJIS, 2021 &VGG16     &41.6  \\ 
    &PLA-DAP\cite{xiong2022source}  &PR, 2022 &VGG16     &42.2  \\ 
    &MT                             &NIPS, 2017 &VGG16     &40.8     \\ 
    &MT+VAT                         &TPAMI, 2018 &VGG16     &42.7     \\ 
    &IRG\cite{vibashan2023instance} &CVPR, 2023 &ResNet50  &45.2 \\ 
    &MemCLR\cite{vs2023towards}     &WACV, 2023 &ResNet50  &44.2  \\ 
    &LPU\cite{chen2023exploiting}   &ACM MM, 2023 &VGG16     &\underline{47.3}  \\ \midrule
    SFOD &Ours                      &This work &VGG16     &\textbf{47.5} \\ 
    \bottomrule 
    \end{tabular}}
    \end{table}
    
\textit{3)} \textit{Results on Sim10k} $\rightarrow$ \textit{Cityscapes}: Table \ref{table.4} compares the domain adaptation results of our methods at natural imaging perspectives. The proposed method achieved 47.5\% AP from sim10k to the car class of cityscapes, which is higher than the latest SFOD methods such as LPU (47.3 \%AP) and MemCLR (44.2 \%AP). Although our method is proposed based on the characteristics of RS images, our method also shows competitiveness on commonly used data sets in the field of computer vision. This shows that our method can not only be used in RS scenarios but can also be generalized to more fields.

\subsection{Further Empirical Analysis}
The proposed method is based on MT and includes three core modules: AFSP, MSP, and PFD. Below, we will analyze the effectiveness and optimal settings of each module through ablation experiments.

    \begin{table}[t]\caption{Ablation study on synthetic-to-real adaptation.}
    \label{table.5}
    \centering
    \resizebox{\linewidth}{!}{
    \begin{tabular}{c|ccc|c|c|c}
    \toprule
   & AFSP &MSP &PFD             & UCAS-AOD  &DLR-3K     &ITCVD  \\ \midrule
   MT &  &  &                      & 61.6      & 65.7      &52.5   \\  
   w/ AFSP & \checkmark & &             & 71.7      &71.3       &66.3   \\ 
   w/ MSP &  &\checkmark &  & 72.5      &70.6       &70.1   \\ 
   w/o PFD  &\checkmark &\checkmark &   & 73.3      &71.5       &70.4   \\  \midrule
   Ours & \checkmark &\checkmark &\checkmark &73.6  &72.4    &72.3   \\ 
    \bottomrule 
    \end{tabular}}
    \end{table}

    \begin{table}[t]\caption{Ablation study on cross-sensor adaptation.}
    \label{table.6}
    \centering
     \resizebox{\linewidth}{!}{
    \begin{tabular}{c|ccc|c|c|c}
    \toprule
   & AFSP &MSP &PFD             & DIOR  &UCAS-AOD     &NWPU VHR-10  \\ \midrule
   MT &  &  &                      & 65.9      & 73.5     &55.0   \\  
   w/ AFSP & \checkmark & &             & 67.2      &75.0       &59.2   \\ 
   w/o AFSP  & &\checkmark & \checkmark  & 71.7      &75.8       &59.2   \\ \midrule
   Ours & \checkmark &\checkmark &\checkmark &72.6  &76.2    &60.7   \\ 
    \bottomrule 
    \end{tabular}}
    \end{table}
    
\textit{1) Analysis of Each Module}: Table \ref{table.5} and \ref{table.6} respectively present the ablation results of three modules in six domain adaptation experiments. As shown in Table \ref{table.5}, compared with MT, only adding a perturbation module AFSP or MSP can significantly improve the model performance. Specifically, the MSP module brings higher benefits than the AFSP module on UCAS-AOD and ITCVD, while the opposite is true on DLR-3K. Compared with a single perturbation module, a small performance improvement can be obtained by comprehensively utilizing these two perturbation modules. Comparing w/o PFD and our method it can be seen that PFD is also an important effective module. From Table \ref{table.6}, we can obtain a similar conclusion. The comprehensive use of three modules achieves the best results in all experiments.

    \begin{table}[t]\caption{Ablation study on the AFSP module design.}
    \label{table.7}
    \centering
     \resizebox{\linewidth}{!}{
    \begin{tabular}{c|cc|c|c|c}
    \toprule
  Method & Adversarial &Style                   & UCAS-AOD  &DLR-3K     &ITCVD  \\ \midrule
  MT  &   &                                     & 61.6      & 65.7      &52.5    \\  
  AFP  & \checkmark &                           & 66.4	  & 67.7	  &63.7   \\ 
  RFSP  &  & \checkmark     & 71.3      & 70.7      &65.3   \\ 
  AFSP(ours)  & \checkmark &\checkmark          & 71.7      & 71.3      &66.3   \\ 
    \bottomrule 
    \end{tabular}}
    \end{table}

\textit{2) Analysis of the AFSP Module Design}: The proposed AFSP module includes two characteristics: adversarial perturbation(rather than random perturbation) and channel-wise statistical perturbation (rather than perturbing feature element). To analyze the necessity of these two characteristics, we designed two modules for comparison: a random feature style perturbation and an adversarial feature perturbation, denoted as RFSP and AFP, respectively. We compared these three designs by adding a single module based on MT. As shown in Table \ref{table.7}, the performance of the RFSP module is slightly lower than our AFSP module, while the AFP module has a significant difference compared to the other two designs. This indicates that style difference is an important domain gap between the source domain and the target domain.

 \begin{figure*}[tbp]    
      \centering           
      \subfloat[Magnitudes of AFSP ablation study.]   
      {
          \label{fig:10subfig1}\includegraphics[width=0.32\textwidth]{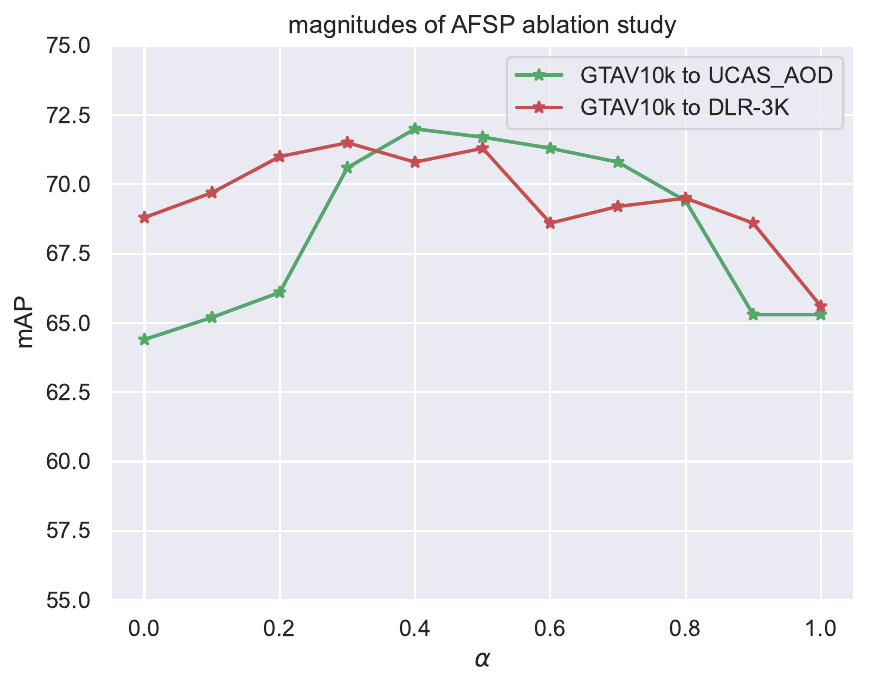}
      }
      \subfloat[Thresholds of pseudo-labeling ablation study.]
      {
          \label{fig:10subfig2}\includegraphics[width=0.32\textwidth]{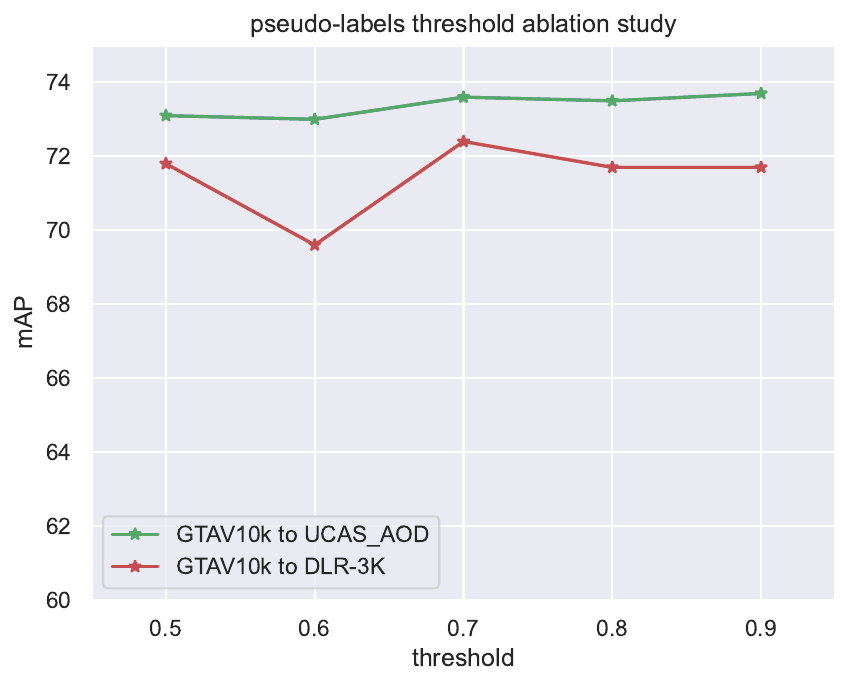}
      }
      \subfloat[Update parameter of EMA ablation study.]
      {
          \label{fig:10subfig3}\includegraphics[width=0.32\textwidth]{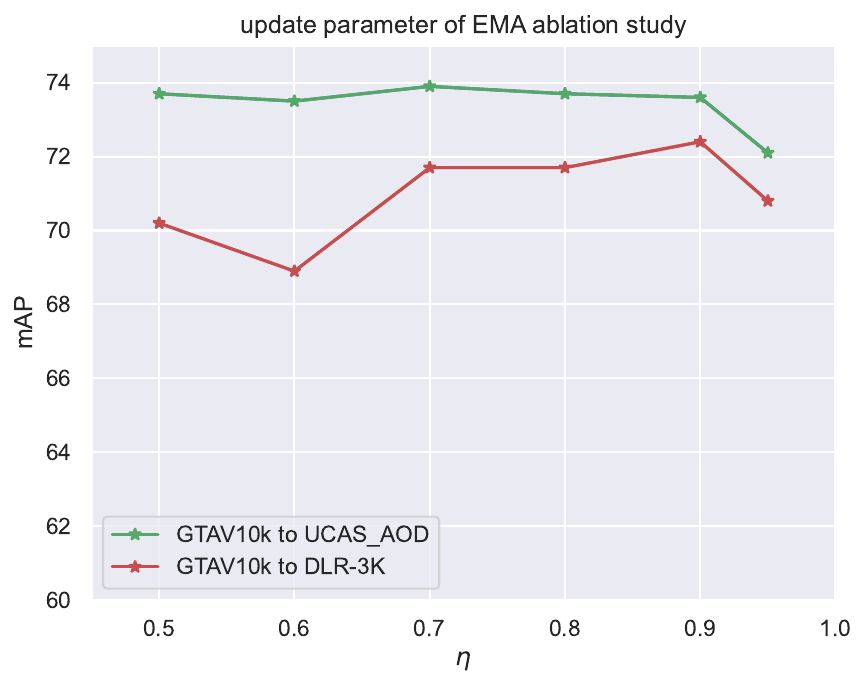}
      }
      \caption{Ablation study of three hyperparameters sensitivity. The amplitude experiment of AFSP ablation is based on the MT framework, without the MSP module and the PFD module. (b) and (c) are based on our method. }   
      \label{fig:10}       
    \end{figure*}

\textit{3) Analysis of Hyperparameter Sensitivity}: We analyze three hyperparameters in the proposed methods, namely the perturbation amplitude of AFSP, the threshold $\tau$ in pseudo-labels, and the update parameter $\eta$ of EMA. In the perturbation method, an important hyperparameter is the setting of the perturbation amplitude. A perturbation amplitude that is too small cannot increase sufficient perturbation changes, while a perturbation that is too large affects the convergence of training. Since the AFSP module features in this article are scaled by the L1-norm and mixed with the input features before output, the perturbation amplitude is determined by the mixing coefficient $\alpha$. We carried out the perturbation amplitude experiment of AFSP based on MT. As shown in Fig. \ref{fig:10} (left), in the GTAV10k $\rightarrow$ UCAS-AOD and GTAV10k $\rightarrow$ DLR-3K experiments, the model performance is the highest when $\alpha$ is between 0.3 and 0.5. Therefore we set $\alpha$ to 0.5 by default. We only change one hyperparameter at a time on the proposed method to analyze  the threshold $\tau$ and the update parameter $\eta$. Fig. \ref{fig:10} (middle) shows the impact of the pseudo label threshold, and Fig. \ref{fig:10} (right) shows the impact of the EMA update parameters. It can be seen that the method in this article is not sensitive to these two thresholds.

\section{Limitations}
The limitations of our methods mainly include the following two parts:
\begin{itemize}
\item 
The proposed perturbation method aims to suppress domain-variant features and make the model focus on domain-invariant features, which improves the cross-domain transferability. However, this may affect the feature discriminability in the target domain. How to balance the transferability and discriminability is a challenge in domain adaptation \cite{chen2020harmonizing}. One possible strategy is to learn discriminative features through self-supervised learning.
\item 
This article mainly focuses on the color, noise, and image style attributes in RS image domain shifts, without exploring geometric issues such as instance scale and rotation. Although the RoIAlign layer samples different targets to the same size, it still cannot fully cope with the scale variation in the target domain. Future work can try multi-scale learning techniques to alleviate this problem.
\end{itemize}

\section{Conclusion}
This article proposes a new source-free domain adaptive object detection method for RS images based on  perturbed domain generation and alignment, which is the first research in this field. We first introduced the source-free object detection setting for RS images. Then two levels of perturbation modules and prototype feature distillation modules were proposed. Our method forces the detector to learn domain invariant features by perturbing domain-variant characteristics of the target domain, thereby improving performance in the target domain. Experiments on multiple datasets have demonstrated the superior ability of our method to detect target domain objects without accessing the source domain data. Experiments in the field of computer vision have also shown that our method is worth promoting.

\bibliographystyle{IEEEtran}
\bibliography{ref}
\end{document}